\documentclass[10pt,twocolumn,letterpaper]{article}

\usepackage[]{cvpr}      

\usepackage[dvipsnames]{xcolor}

\usepackage{graphicx}
\usepackage{amsmath}
\usepackage{amssymb}
\usepackage{booktabs}
\usepackage{pifont}
\usepackage{caption}

\usepackage{threeparttable}
\usepackage{multirow}

\usepackage[normalem]{ulem}
\usepackage{marvosym}
\usepackage{soul}

\usepackage{colortbl}
\usepackage{array}
\usepackage{algorithm}
\usepackage{algpseudocode}
\usepackage[utf8]{inputenc}
\usepackage{mathtools}
\usepackage[accsupp]{axessibility}

\usepackage{subcaption}
\usepackage{booktabs} 

\definecolor{cvprblue}{rgb}{0.21,0.49,0.74}
\usepackage[pagebackref,breaklinks,colorlinks,citecolor=cvprblue]{hyperref}

\usepackage[capitalize]{cleveref}
\crefname{section}{Sec.}{Secs.}
\Crefname{section}{Section}{Sections}
\Crefname{table}{Table}{Tables}
\crefname{table}{Tab.}{Tabs.}

\title{Audio-Visual Segmentation via Unlabeled Frame Exploitation}

\author{Jinxiang Liu$^{1}$, \ Yikun  Liu$^{1}$, \ Fei Zhang$^{1}$, \ Chen Ju$^1$,  Ya Zhang$^{1,2} \textsuperscript{\Letter}$, \ Yanfeng Wang$^{1,2}$\\[3pt]
$^1$ Cooperative Medianet Innovation Center, Shanghai Jiao Tong University \ \
$^2$ Shanghai AI Laboratory\\
{\tt\small \{jinxliu,\,yikunliu,\,ferenas,\,ju\_chen, ya\_zhang,\,wangyanfeng622\}@sjtu.edu.cn}
}

\begin{document}

\twocolumn[{%
\renewcommand\twocolumn[1][]{#1}%
\maketitle
\vspace{-0.5cm}
\begin{center}
    \centering
    \includegraphics[width=1.0\linewidth, trim=0 18 0 0]{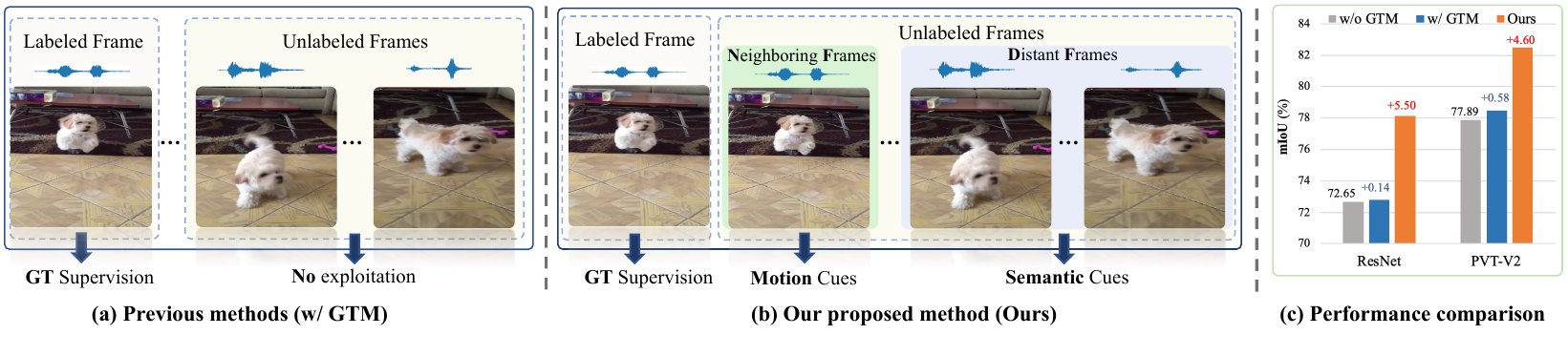}
    \captionof{figure}{
   \textbf{Comparison between previous methods and ours on how to harness the unlabeled frames. } (a) Previous methods perform global temporal modeling (GTM) to process all frames from a sequence including labeled and unlabeled ones, without the exploitation of the unlabeled frames.  (b) Our method employs two types of unlabeled frames: (i) the \textit{neighboring frames} (\textbf{NFs}) provide motion cues for accurately segmenting the sounding object; (ii) the \textit{distant frames} (\textbf{DFs}) contain semantic cues for enhancing data diversity. (c) Based on TPAVI method, compared to the model trained only using labeled frames (w/o GTM), previous methods using global temporal modeling (w/ GTM) only show marginal performance gain; while our method achieves significant improvement with the unlabeled frames.
    }
\label{fig:teaser}
\end{center}%
}]

\begin{abstract}
\vspace{-0.5cm}
Audio-visual segmentation (AVS) aims to segment the sounding objects in video frames. Although great progress has been witnessed, we experimentally reveal that current methods reach marginal performance gain within the use of the unlabeled frames, leading to the underutilization issue. To fully explore the potential of the unlabeled frames for AVS, we explicitly divide them into two categories based on their temporal characteristics, i.e., neighboring frame (NF) and distant frame (DF). NFs, temporally adjacent to the labeled frame, often contain rich motion information that assists in the accurate localization of sounding objects. Contrary to NFs, DFs have long temporal distances from the labeled frame, which share semantic-similar objects with appearance variations. Considering their unique characteristics, we propose a versatile framework that effectively leverages them to tackle AVS. Specifically, for NFs, we exploit the motion cues as the dynamic guidance to improve the objectness localization. Besides, we exploit the semantic cues in DFs by treating them as valid augmentations to the labeled frames, which are then used to enrich data diversity in a self-training manner. Extensive experimental results demonstrate the versatility and superiority of our method, unleashing the power of the abundant unlabeled frames.

\end{abstract}

\section{Introduction}\label{sec:intro}
Humans perceive the surroundings not only by seeing but also by hearing to accurately and efficiently obtain the target information~\cite{guttman2005hearing}.
In the audio-visual understanding field, the demand to visually attend to the auditory objects has driven the exploration of the audio-visual segmentation (AVS) task~\cite{zhou2022avs}.
The goal of AVS is to localize and segment the sounding objects in the video frames with the guidance of audio signals.
And the successful grounding of auditory objects with the AVS task will benefit a wide range of downstream tasks such as multi-modal content editing~\cite{agarwal2023audio,lee2022sound,lee2022sound1}, video surveillance~\cite{crocco2016audio,cristani2007audio}, and robot industry~\cite{gan2020look,wu2009surveillance}.

To address the task, current methods~\cite{zhou2022avs,Mao_2023_ICCV,gao2023avsegformer,li2023catr,liu2024annotation,liu2023audio,liu2023audio1} are based on the dataset which is \textit{sparsely} annotated.
Concretely, due to the high labeling costs, only few frames in a video frame sequence are annotated with groundtruth masks, leaving the rest abundance of frames unlabeled.
For example, in AVSBench-S4 dataset~\cite{zhou2022avs}, only one sampled frame is annotated for a 5-second video.
Despite the predominance of unlabeled frames within the datasets, current approaches~\cite{zhou2022avs,Mao_2023_ICCV,gao2023avsegformer,li2023catr,liu2023audio,liu2023audio1} adopt global temporal modeling module (GTM) that overemphasizes on exploiting the labeled frames to help address AVS, which may lead to the \textit{underutilization} of the abundant unlabeled frames.
To further verify this, we perform experiments based on the typical TPAVI~\cite{zhou2022avs} method. 
The results in~\cref{fig:teaser} (c) demonstrate that compared with the baseline model (w/o GTM) trained with only labeled frames, current approach (w/ GTM) without tailored handling for the unlabeled frames only provides marginal improvement.
Therefore, we are motivated to explore a more effective way to utilize the unlabeled frames for the AVS task.

Before delving into the exploitation of the unlabeled frames, let us rethink the characteristics of the abundant unlabeled frames.
Taking~\cref{fig:teaser} (b) as an example, given a target labeled frame describing ``dog jumping'', its neighboring unlabeled frames usually have very tiny visual appearance changes.
For the distant frames, they usually contain the same object but with large appearance variations to the object in the labeled frame, \eg, the dog has transformed from the pose ``jumping" in the labeled frame to ``walking" or ``standing still'' in the distant frames.
Based on the observation, we start by first dividing the unlabeled frames into two categories: \textit{neighboring frame} (NF) and \textit{distant frame} (DF), based on the temporal distance with the target labeled frame. 
Though the visual changes are very limited, NFs often contain rich dynamic motion information that is important to the audio-visual understanding~\cite{NIPS2000_11f524c3,EphratMLDWHFR18,ZhaoGM019,singh2023flowgrad,Barzelay2007}.
If properly used, the motion can not only assist in the accurate localization of the sounding objects but also provide the shape details of objects.
For the DFs, both they and the labeled frame reflect the different stages of an audio-visual event ~\cite{mahmud2023ave,guo2022look,guttman2005hearing,shvets2019leveraging}.
Contrary to the NFs, this long-term temporal relationship means that the DFs generally share the same or semantic-similar objects but with large appearance variations.
Therefore, DFs could serve as the natural \textit{semantic augmentations} for the labeled frames, which can be utilized to diversify the training data, thereby enhancing the model generalization capabilities.

Considering the characteristics of NFs and DFs, we propose a universal \textit{unlabeled frame exploitation (UFE)} framework to leverage the two types of unlabeled frames with different strategies.
For NFs, we extract the motion by calculating the optical {flow} between the target labeled frame and its NFs.
And we explicitly feed the flow as model input to incorporate the motion guidance, which is complementary to the still RGB frame.
In terms of DFs, since they are the natural \textit{semantic augmentations} to labeled frames, the training data could be significantly enriched beyond the labeled frames.
To this end, we propose a teacher-student network training framework to provide valid supervision for the unlabeled frames with the \textit{weak-to-strong consistency}, where the predictions for the strong-augmented frames from the student are supervised by the predictions for the weak-augmented ones from the teacher.
We perform the experiments by applying our proposed framework to two representative methods TPAVI~\cite{zhou2022avs} and AVSegFormer~\cite{gao2023avsegformer}.
Extensive experimental results demonstrate the effectiveness of our proposed method to attack the AVS task by exploiting the unlabeled frames.
The main contributions are:

\begin{itemize}
    \item We propose a simple but effective partition strategy for the unlabeled frames based on the temporal characteristics, i.e., \textit{neighboring frames} and \textit{distant frames}, relieving the \textit{underutilization} issue in AVS.
    
        \item We propose UFE, a versatile framework that leverages the NFs and DFs, where NFs provide motion guidance and DFs enhance the data diversity beyond the labeled frames, explicitly improving the objectness segmentation.  

    \item Extensive experiments show our method can effectively exploit the abundant unlabeled frames and achieves new state-of-the-art performance on the AVS task, \eg., 78.96 mIoU with ResNet backbone and 83.15 mIoU with PVT backbone on AVSBench-S4 dataset.

\end{itemize}

\section{Related Work}\label{sec:related}
\noindent \textbf{Audio-Visual Segmentation.} With the advancement of multi-modal learning~\cite{ju2022prompting,ju2023constraint,ju2023multi,ju2023turbo,cheng2023mixer}, many audio-visual understanding problems have been studied, such as audio-visual sound separation~\cite{zhao2018sound,zhao2019sound,gao2021visualvoice,ephrat2018looking,tzinis2022audioscopev2}, 
audio-visual segmentation~\cite{hu2021class,hu2020discriminative,hu2019deep,lin2021unsupervised,chen2021localizing,song2022self,liu2022exploiting,senocak2018learning,qian2020multiple,liu2024annotation} and audio-visual video understanding~\cite{kazakos2019epic,tian2020unified,lin2019dual,lee2020cross,tian2020unified}.
In this paper, we focus on the audio-visual segmentation (AVS) task, whose purpose is to segment the sounding objects in video frames.
Previous methods~\cite{hu2021class,hu2020discriminative,hu2019deep,lin2021unsupervised,chen2021localizing,song2022self,liu2022exploiting,senocak2018learning,qian2020multiple,owens2018audio} usually tackled the task in self-supervised or weakly-supervised learning and termed the task as visual sound source localization.
Recently researchers~\cite{zhou2022avs,Mao_2023_ICCV,gao2023avsegformer,li2023catr,liu2023audio,liu2023audio1,liu2024annotation} have been tackling AVS under the umbrella of supervised learning on the~\textit{sparsely-annotated} AVSBench dataset~\cite{zhou2022avs}.
And based on architecture, we divide these methods into two categories: FCN-based~\cite{zhou2022avs,Mao_2023_ICCV} and transformer-based methods~\cite{gao2023avsegformer,li2023catr,liu2023audio,liu2023audio1,liu2024annotation}.
For the FCN-based methods, the typical model is TPAVI~\cite{zhou2022avs}.
The key design of ~\cite{zhou2022avs} is the temporal pixel-wise audio-visual interaction (TPAVI) module which performs audio-visual feature fusion similar to the non-local block~\cite{wang2018non}. 
In terms of the transformer-based models~\cite{gao2023avsegformer,li2023catr,liu2023audio,liu2023audio1,liu2024annotation}, the general idea is to achieve audio-visual fusion and mask decoding with transformer~\cite{vaswani2017attention}.
For example, AVSegFormer~\cite{gao2023avsegformer} employs cross-attention to merge the spatial-temporal mask visual features and the audio embeddings.
Although the architecture designs of FCN-based and transformer-based methods differ, they share a similar architecture pipeline, which consists of feature extraction, audio-visual fusion, and mask decoding.
Besides, we observe current methods usually process the labeled frames and unlabeled frames from a video equally and predict the segmentations for all frames.
However, due to the lack of annotations for the unlabeled data, the predictions for the unlabeled frames have \textit{no} supervision.
Inevitably, the methods without special handling for unlabeled frames lead to a suboptimal utilization problem.

\noindent \textbf{Motion and Sound.} Object motions and air vibration cause sound.  
We humans usually perceive sound together with the motion of visual objects.
This strong relation between sound and motion has been studied by previous methods, \eg, \cite{Kidron1467253,franccoise2014probabilistic} employed probabilistic models to investigate the relationship between motion and sound.
Moreover, many previous works~\cite{NIPS2000_11f524c3,gabbay2018seeing,ChungSVZ17,EphratMLDWHFR18,OwensE18,ZhaoGM019,fedorishin2023hear,singh2023flowgrad,Barzelay2007} have shown the important role that motion plays in audio-visual learning.
For example, lip motion is a vital clue for speech processing tasks such as speech denoising~\cite{gabbay2018seeing} and speech separation~\cite{EphratMLDWHFR18,OwensE18}.
Other studies~\cite{Barzelay2007,ZhaoGM019} also modelled the temporal motion information~\cite{zhao2020bottom,ju2021divide,ju2022adaptive,ju2020point} in visual frames to solve the cocktail-party problem.
However, few works explore motion for the AVS task.

\noindent \textbf{Teacher-Student Network.} Teacher-student network has become the dominant architecture for many problems~\cite{sohn2020fixmatch,hinton2015distilling,gou2021knowledge,french2019semi,hu2021semi, zou2020pseudoseg,yang2023revisiting,liu2021unbiased,tang2021humble,ju2023distilling,xu2021end}.
In a teacher-student network, the prediction of the teacher model is used to regularize the prediction of the student model, thereby, transferring the teacher's knowledge.
When employed to handle scenarios encompassing both labeled and unlabeled data, the teacher trained with labeled data can generate pseudo-labels for the unlabeled data, which the student then tries to match.
In our model design, we first separate the labeled frames and unlabeled frames from a sequence for independent processing, and frame-wisely consider the labeled and the unlabeled frames across the dataset in our framework.
To exploit the unlabeled frames, we adopt the weak-to-strong consistency~\cite{sohn2020fixmatch,yang2023revisiting,melas2021pixmatch,liu2021unbiased} by regularizing the predictions for the strong-augmented unlabeled frames from the student with the predictions for the weak-augmented unlabeled frames from the teacher.
Thereby, the unlabeled frames can obtain supervision in a self-supervised manner.

\section{Motivation}\label{sec:motivation}
For the task of audio-visual segmentation (AVS), the input data consists of a sequence of sampled video frames $\mathcal{V}=\left\{I_i\right\}_{i=1}^T$, where $I_{i} \in \mathbb{R}^{3 \times H_{0} \times W_{0}}$, and its corresponding audios $\mathcal{A}=\left\{a_i\right\}_{i=1}^T$, where $a_{i} \in \mathbb{R}^{d}$ is the audio clip with each video frame.   
The objective of the AVS task is to segment the sounding objects corresponding to the its audio in each frame in $\mathcal{V}$.
The target segmentation is the binary masks for each frame $\mathcal{Y}=\left\{y_i\right\}_{i=1}^T$, where $y_{i} \in  \{0,1\}^{{H_{0} \times W_{0}}}$.

As mentioned in~\cref{sec:related}, current mainstream approaches to AVS falls into two categories: the FCN-based methods~\cite{zhou2022avs,Mao_2023_ICCV} and transformer-based methods~\cite{gao2023avsegformer,li2023catr,liu2023audio,liu2023audio1,liu2024annotation}.
Though the methods vary, the pipelines of methods can be abstracted into three subsequent steps: 
feature extraction with image encoder $\Phi_{\text{image}}$ and audio encoder $\Phi_{\text{audio}}$, multi-modal feature fusion with fusion module $\Phi_{\text{fusion}}$, and mask prediction with mask decoder $\Phi_{\text{dec}}$.
Formally, the predicted segmentation $\mathcal{P}$ is obtained with:

\begin{equation}
    \mathcal{P} =\Phi_{\text{dec}}\left(\Phi_{\text{fuse}}\left(\Phi_{\text{image}}\left( \mathcal{V} \right), \Phi_{\text{audio}}\left(\mathcal{A}\right)\right)\right).
\end{equation}

Notably, mainstream methods treat the labeled frames and unlabeled frames sampled from a video sequence equally and predict the masks for all frames.
However, only the labeled frames have groundtruth supervision while the remaining abundant unlabeled frames have \textit{no} supervision.
And the only possible benefit which the unlabeled frames might provide for labeled frames is the contextual information with the global temporal modeling (GTM) operation.
Concretely, global temporal modeling (GTM) employs cross-attention~\cite{vaswani2017attention} to model the temporal relationships of the features across all the frames from a video, including labeled and unlabeled ones.
To illustrate, \citet{zhou2022avs} deployed the cross-attention to integrate the space-time relations of the features in the TPAVI module in the audio-visual fusion stage; likewise, \citet{gao2023avsegformer} proposed the channel-attention mixer based on the cross-attention in the audio-visual fusion stage to obtain the mask features.

To measure the improvement by exploiting the unlabeled frames with GTM of the previous method~\cite{zhou2022avs,gao2023avsegformer}, we establish the baseline by discarding the unlabeled frames and only using the labeled frames for model training.
We perform experiments on AVSBench-S4 dataset and compare the performance with two typical methods TPAVI~\cite{zhou2022avs} and AVSegFormer~\cite{gao2023avsegformer}.
The results on TPAVI baseline model are shown in~\cref{fig:teaser} (c), compared to the model trained with only labeled frames (w/o GTM), previous method~\cite{zhou2022avs} based on  global temporal modeling (w/ GTM) achieves only marginal performance gain: 0.14 gain with ResNet and 0.58 gain with PVT in mIoU ($\mathcal{M}_{\mathcal{J}}$).
For more metrics and more results on the AVSegFormer baseline method, please refer to the supplementary materials.

The results demonstrate the major issue of current methods: the  \textit{underutilization} of the unlabeled frames to boost the performance for the AVS task.
Based on the observation, we intend to devise a more effective method to fully exploit the unlabeled frames, which is elaborated as follows.

\begin{figure*}[hbt]
\centering
\includegraphics[width=0.97\linewidth]{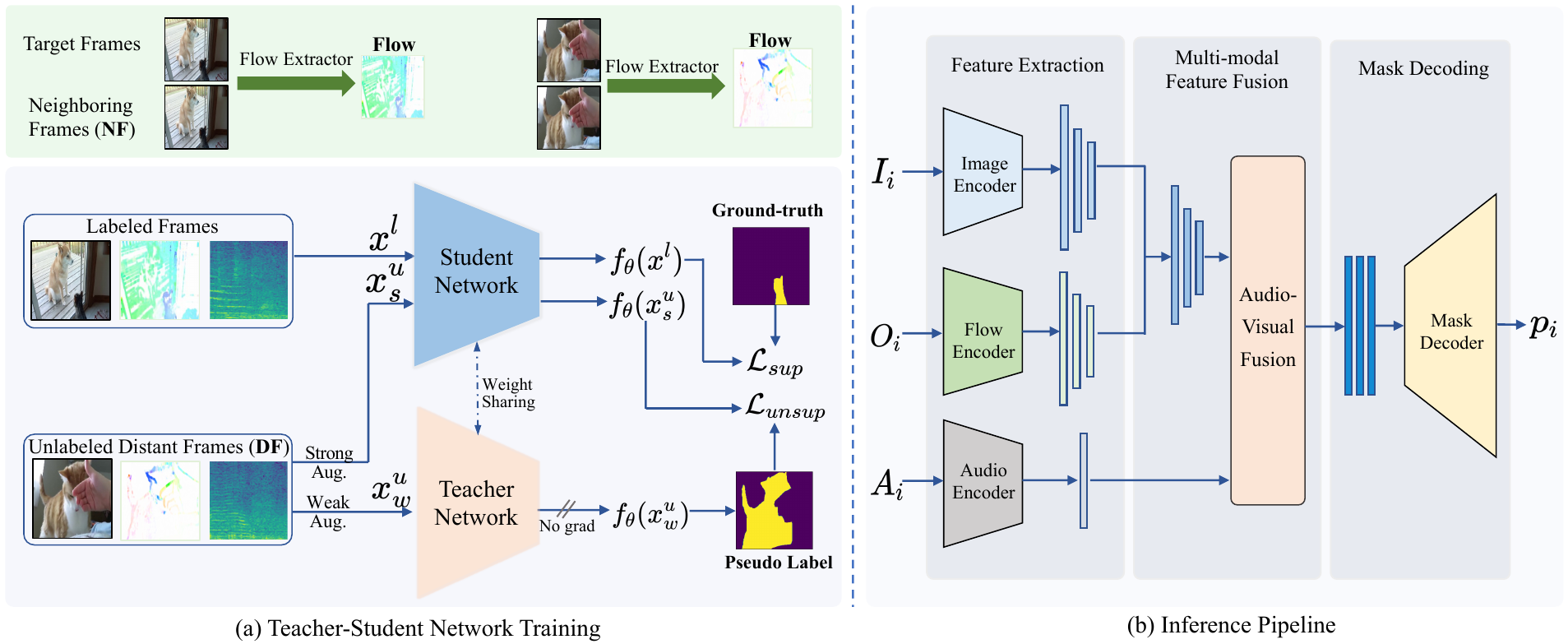}
\caption{Overview of our framework to exploit unlabeled frames.
(a) \textbf{Teacher-student network for training}. Student network is optimized with $\mathcal{L}_{sup}$ and $\mathcal{L}_{unsup}$. 
$\mathcal{L}_{sup}$ is computed with the predicted mask $f_\theta(x^l)$ and its groundtruth for the labeled frame $x^l$;
$\mathcal{L}_{unsup}$ is computed between $f_\theta(x^u_s)$ from the student and the predicted pseudo mask $f_\theta(x^u_w)$ for the strong-augmented unlabeled image from teacher. 
(b) \textbf{Inference pipeline} of the framework. We incorporate flow as auxiliary input to exploit the motion cues within NFs.
}
\vspace{-0.5cm}
\label{fig:framework}
\end{figure*}

\section{Method}\label{sec:method}
\subsection{Framework Overview}
Technically, our proposed framework can be established based on either FCN-based methods or transformer-based methods.
Since both categories of methods are divided into three steps as elaborated in~\cref{sec:related}, our proposed framework also has three steps in the inference stage as illustrated in~\cref{fig:framework} (b).
The model accepts the image $I_i$, the calculated flow $O_i$ and its audio $A_i$ as model inputs, then goes through three successive steps, to predict the target segmentation mask.
The main difference from previous methods is that we harness the optical flow $O_i$ extracted from the target frame and its unlabeled \textit{neighboring frame} (NF) as model input, in order to exploit the motion information to guide the model to focus on the exact sounding object thereby achieving accurate segmentation.

Regarding the exploitation of \textit{distant frames} (DFs), we consider leveraging the semantic information of the distant frames and integrating the abundance of DFs to enrich the training data diversity.
Even though DFs have \textit{no} groundtruth annotations, we adopt the teacher-student network in the training phase, as shown in~\cref{fig:framework} (a).
Different from previous methods where the unlabeled frames receive \textit{no} supervision, our teacher-student network can provide supervision for the unlabeled distant frames to exploit the unlabeled frames with the \textit{weak-to-strong consistency} from pseudo-labeling.

\subsection{Neighboring Frame Exploitation}
Sound occurs due to object motions and air vibration, thus sound has strong associations with motions.
For example, when someone speaks, it always comes along with the movement of the speaker's lips; the sound of a musical instrument usually comes along with the hand movement of the player.
The motion information can provide important dynamic cues for achieving the AVS task, which is complementary to the static information that still RGB frame provides.
Concretely, i). motion can assist in localizing the exact sounding object and resolve the identity ambiguity.
ii). motion information can even outline the shape and contour of the sounding objects, which contributes to accurate segmentations with better fine-grained details.  
And the motion information of the target frame can be simply extracted by exploiting the neighboring unlabeled frames. 
Concretely, we use the the target frame and its temporally-adjacent unlabeled frame in raw frame sequence to compute the optical flow, which serves a common way for motion estimation~\cite{simonyan2014two,fortun2015optical}.
And we leverage the optical flow by incorporating it as model input together with the target frame.

Specifically, for the $i$th sampled target frame from a video to be sent into the model, we utilize the RGB frame $I_i^l$ and its subsequent neighboring unlabeled frame $I_i^{n}$ to calculate the optical flow $O_i$
with Gunnar Farneback algorithm~\cite{farneback2003two}.
Then as shown in~\cref{fig:framework} (b) we extract image and optical flow features with the image encoder and optical flow encoder separately. 
Following \cite{zhou2022avs, gao2023avsegformer}, the image encoder can be either ResNet50~\cite{he2016deep} or PVT-v2~\cite{wang2022pvt}.
We represent the extracted multi-scale image features as $\mathcal{F}_{vi} \in \mathbb{R}^{h_i \times w_i \times C_i}$, where $(h_i, w_i) = (H, W) / 2^{i + 1}$ for $i = 1, ..., 4$. 
For the optical flow encoder, we adapt the pretrained ResNet-18 architecture by modifying the in-channel number of the first convolution layer into 2. 
We denote the extracted optical flow features from the flow encoder as $\mathcal{F}_{fi} \in \mathbb{R}^{h_o \times w_o \times C_o}$.
And for the audio encoder,  we adopt the VGGish \cite{hershey2017cnn} model pretrained on the AudioSet~\cite{gemmeke2017audio} dataset and pool the features into a vector embedding.
The obtained audio features are denoted as
$\mathcal{F}_{a} \in \mathbb{R}^{C_a}$.

Before performing audio-visual fusion, we first employ a refinement network to fuse the extracted multi-scale image features and flow features.
We first utilize the upsampling operations to ensure the refined flow features align with the visual features. 
Then we fuse the refined optical flow features with the visual features of scale with summation.
This process can be formulated as:
\begin{equation}
    \mathcal{F}_{\text{$refine_i$}} = \mathcal{F}_\text{vi} +  \Phi_\text{Upsample}(\Phi_{\text{Refine}}(\mathcal{F}_{fi})),
\end{equation}
where $\Phi_{\text{Refine}}$ is composed of multiple convolution layers. 

After fusing the image and optical flow features, we obtain the aggregated visual features  $ \mathcal{F}_{\text{$refine_i$}}$; then we perform the multi-modal feature  fusion and mask decoding to obtain the segmentation predictions, by inheriting the modules from TPAVI~\cite{zhou2022avs} or AVSegFormer~\cite{gao2023avsegformer}.

\subsection{Distant Frame Exploitation}
\textit{Distant frames} (DFs) refer to the video frames which are temporally faraway from the target labeled frame. 
Contrary to neighboring frames, the distant frames do not contain the motion dynamics of the target sounding objects due to the long temporal distance.
However, thanks to the large visual appearance variations to the target labeled frame, these distant frames are the natural \textit{augmentations} to the target labeled frames with shared semantics.
And these frames can substantially enhance the data diversity if utilized for model training, thereby boosting the model generalization.
Even though there are no groundtruth annotations for the DFs, modern pseudo-labeling techniques can be harnessed to provide self-supervision.
Given the observation, we propose to exploit the unlabeled distant frames with a teacher-student network to train the model, inspired by recent works~\cite{sohn2020fixmatch,liu2021unbiased,liu2022unbiased,mi2022active,sun2023refteacher}.

Specifically, given the training data, we first divide it into the labeled frame set and the unlabeled distant frame set.
We first use the labeled frames to train the teacher for some iterations to ensure that the model has the capability to generate reliable pseudo mask labels; this step is called burn-in stage~\cite{liu2021unbiased,liu2022unbiased,mi2022active,sun2023refteacher}.
Then we initialize the student network with the weights of the teacher; and during the training stage, the teacher and student network share the weights.
The difference is that the model parameters are optimized through the student network with the supervised loss and unsupervised loss; while the teacher network with \texttt{stop\_gradient} serves to provide pseudo labels for the unlabeled frames.
To this end, in each iteration after the burn-in stage, the student network is optimized with labeled frames \{$(I_{\ell}, O_{\ell}, A_{\ell}), y_{\ell} $\} and the unlabeled frames  \{$(I_u, O_u,A_u) $\}.
For the labeled frames, the supervised loss $\mathcal{L}_\text{sup}$ is computed between student prediction and groundtruth mask. 
For the supervised loss $\mathcal{L}_\text{sup}$, it can be either BCE loss~\cite{zhou2022avs} in TPAVI or Dice loss~\cite{milletari2016v} in AVSegFormer~\cite{gao2023avsegformer}, which are formulated as:

\begin{equation}\label{eq:bceloss}
    \mathcal{L}_{BCE} =  -\frac{1}{N} \sum_{i=1}^{N} \left[ y_i \cdot \log(p_i) + (1 - y_i) \cdot \log(1 - p_i) \right],
\end{equation}

\begin{equation}\label{eq:diceloss}
     \mathcal{L}_\text{Dice} =  1 - \frac{2 \sum_{i=1}^{N} (p_i \cdot y_i)}{\sum_{i=1}^{N} p_i + \sum_{i=1}^{N} y_i},
\end{equation}
where in~\cref{eq:bceloss} and~\cref{eq:diceloss}, $y_i$ represents the ground truth label of a pixel and $p_i$ represents the predicted probability of a pixel belonging to the foreground class.

For the unlabeled frames \{$(I_u, O_u,A_u) $\}, the input visual signals are perturbed by two operators, i.e., weak perturbation $\mathcal{H}^w$ (\eg, flip) and strong perturbation $\mathcal{H}^s$ (\eg, cutmix).
Afterwards, we feed the weakly-augmented view  \{$\mathcal{H}^w(I_u), \mathcal{H}^w(O_u),A_u $\} to the teacher model to predict pseudo mask label $p^w$; and we feed the strongly-augmented view \{$\mathcal{H}^s(I_u), \mathcal{H}^s(O_u),A_u $\} to the student model to predict the mask $p^s$.
The unsupervised loss $\mathcal{L}_\text{unsup}$ ensures that the predictions under strong perturbations align with those under weak perturbations, which can be formulated as:

\begin{equation}\label{eq:unsuploss}
    \mathcal{L}_\text{unsup} = \frac{1}{B_u}\sum H(p^w, p^s),
\end{equation}

\noindent where $B_u$ is the batch size for unlabeled data. $H$ serves to minimize the entropy between two probability distributions:

\begin{equation}
\begin{aligned}
    & p^w = \Phi_{\mathrm{mask}}(H^w(I_i), H^w(O_i), A_i), \\
    & p^s = \Phi_{\mathrm{mask}}(H^s(I_i), H^s(O_i), A_i).
\end{aligned}
\end{equation}

The overall training objective $\mathcal{L}_{\text{total}}$  is a combination of supervised loss $\mathcal{L}_{sup}$ and unsupervised loss $\mathcal{L}_\text{unsup}$ as:

\begin{equation}\label{eq:totalloss}
    \mathcal{L}_{\text{total}} = \mathcal{L}_\text{sup} + \lambda \mathcal{L}_\text{unsup},
\end{equation}
where $\lambda$ is the weight to balance the losses.

\begin{algorithm}
\caption{Algorithm of Our Framework}
\begin{algorithmic}[1]
\Require Labeled frames $\mathcal{D}^L = \{(\mathcal{I}_{\ell}, \mathcal{A}_\ell), \mathcal{Y}_\ell\}$ and its unlabeled neighboring frames $\mathcal{D}^L_{n} = \{\mathcal{I}^\ell_{n}\}$, unlabeled distant frames $\mathcal{D}^{D} = \{\mathcal{I}_{d},\mathcal{A}_d\}$ and its unlabeled neighboring frames $\mathcal{D}^D_{n} = \{\mathcal{I}^d_{n}\}$, burn-in iteration $k$, maximum iteration ${N}$
\Ensure Teacher (Student) Model Weights $\theta^i$
\For{$i < N$}
 \State Sample labeled data from $\mathcal{D}^L$ and its NFs from $\mathcal{D}^L_n$
 \State Calculate the motion flow $\{O_\ell \}$
   \State Compute $L_{sup}$ with~\cref{eq:bceloss} or~\cref{eq:diceloss} 

    \If{$i < k$}
        \State Update $\theta^i$ with $\mathcal{L}_{sup}$ 
    \EndIf
    \If{$i >= k$}
       
    \State Sample unlabeled data from $\mathcal{D}^D$ and its NFs from $\mathcal{D}_n^D$
    \State Calculate the motion flow $\{O_d \}$
        \State Compute $\mathcal{L}_{unsup}$ with~\cref{eq:unsuploss} 
        \State Update $\theta^i$ with $\mathcal{L}_{{total}}$ by~\cref{eq:totalloss}
    \EndIf
\EndFor
\State \Return $\theta^i$
\end{algorithmic}
\end{algorithm}

\section{Experiment}\label{sec:exper}
\subsection{Experimental setup}
\noindent \textbf{Datasets.}
We use the AVSBench~\cite{zhou2022avs} dataset, which was recently proposed for audio-visual segmentation task with segmentation mask annotations for sounding objects.
This dataset has two subsets: the semi-supervised Single Sound Source Segmentation (S4) and the fully supervised Multiple Sound Source Segmentation (MS3).
In S4 subset, there exists only one sounding object in the video. 
For each training video sample, only the first frame of the frame sequence is annotated while all five sampled frames need to be segmented in the validation and test sets.
In MS3 subset, there might exist more than one sounding objects in the video frames.
All five frames sampled from a 5s-long video are provided with mask annotations in both training and evaluation stages.
In terms of the dataset size, S4 subset includes 3452$/$740$/$740 videos in training, validation and test sets separately, for a total of 10,852 annotated frames;
MS3 subset includes 296$/$64$/$64 videos in training, validation and test sets, with 2,120 annotated frames.
For the MS3 dataset, since all five frames are annotated in training set, we extract semantically relevant videos from the VGGSound dataset and select the middle frames as the source of distant frames for the MS3 dataset, totaling 12,990 in size.

\vspace{0.1cm}
\noindent \textbf{Metrics.}
We adopt the mean Intersection-over-Union ($\mathcal{M}_{\mathcal{J}}$) and F-score ($\mathcal{M}_{\mathcal{F}}$) as our evaluation metrics following previous methods~\cite{zhou2022avs,gao2023avsegformer}.

\vspace{0.1cm}
\noindent \textbf{Implementation details.}
Technically, our proposed framework can be combined with any mainstream methods.
In our experiments, we verify our method based on TPAVI~\cite{zhou2022avs} and AVSegFormer~\cite{gao2023avsegformer}, thus we follow their experimental settings such as backbone, learning rate and optimizing strategy.
The input image and optical flow size is $224\times 224$.
For the teacher-student network, the weak augmentations include resize, crop and horizontal flip; and the strong augmentations include additional color jitter, grayscale and cutmix operations. 
And the loss weight $\lambda$ is set to 0.5.
The burn-in stage lasts for 10 epochs.
We train the models for 120 epochs, with one NVIDIA A100 GPU.
Batch size is 24.

\begin{table}[]
\setlength{\tabcolsep}{2.5pt}
\small
    \centering
    \begin{tabular}{lllll}
        \toprule
        \multirow{2}{*}{Method} & \multicolumn{2}{c}{S4} & \multicolumn{2}{c}{MS3} \\
        \cmidrule(lr){2-3} \cmidrule(lr){4-5}
         &  $ \mathcal{M}_{\mathcal{J}} $  & $ \mathcal{M}_{\mathcal{F}} $ &  $ \mathcal{M}_{\mathcal{J}} $ & $ \mathcal{M}_{\mathcal{F}} $ \\
        \cmidrule(lr){0-0} \cmidrule(lr){2-3} \cmidrule(lr){4-5}
        $\text{TPAVI}_\text{(ResNet)}$ & 72.79 & .848 & 47.88 & .578  \\
        $\textbf{Ours}_{(\text{ResNet})} $& $\textbf{78.15}{\,({\color{green}\text{+5.36}})}$ & \textbf{.887} & $\textbf{54.08}{\,({\color{green}\text{+6.20}})}$ & \textbf{.616} \\
        \cmidrule(lr){0-0} \cmidrule(lr){2-3} \cmidrule(lr){4-5}
        $\text{TPAVI}_{(\text{PVT})}$  & 78.74 & .879 & 54.00 & .645  \\
         $\textbf{Ours}_{\text{(PVT)}}$  & $\textbf{82.49}{\,({\color{green}\text{+3.75}})} $& \textbf{.912} & $\textbf{59.49}{\,({\color{green}\text{+5.49}})}$ & \textbf{.676} \\
        \cmidrule(lr){0-0} \cmidrule(lr){2-3} \cmidrule(lr){4-5}
        $\text{AVSegFormer}_{(\text{ResNet})} $ & 76.45 & .859 & 49.53 & .628  \\
     $\textbf{Ours}_{(\text{ResNet})} $ & $\textbf{78.96}{\,({\color{green}\text{+2.51}})}$ & \textbf{.875} & $\textbf{55.88}{\,({\color{green}\text{+6.35}})}$ & \textbf{.645} \\
        \cmidrule(lr){0-0} \cmidrule(lr){2-3} \cmidrule(lr){4-5}
       $ \text{AVSegFormer}_{(\text{PVT})} $ & 82.06 & .899 & 58.36 & .693  \\
         $\textbf{Ours}_{(\text{PVT})}$  & $\textbf{83.15}{\,({\color{green}\text{+1.09}})} $ & \textbf{.904} & $\textbf{61.95}{\,({\color{green}\text{+3.59}})}$ & \textbf{.709} \\
        \bottomrule
    \end{tabular}
    \caption{Comparison of our method and baseline methods on AVSBench {S4} and {MS3} subsets .}
    \label{tab:baseline}
    \vspace{-0.5cm}
\end{table}

\subsection{Comparison with Prior Arts}
\noindent \textbf{Improvement of our method over baselines.}
To verify the effectiveness of our method, we choose two typical baseline methods: FCN-based TPAVI~\cite{zhou2022avs} and transformer-based AVSegFormer~\cite{gao2023avsegformer}, and we apply our framework onto the baseline methods.
We compare the performance between the models (\textbf{Ours}) and the baseline models in~\cref{tab:baseline}.

As~\cref{tab:baseline} shows, our method consistently improves the performance significantly on both TPAVI~\cite{zhou2022avs} and AVSegFormer~\cite{gao2023avsegformer}, which indicates the effectiveness and universality of our method.
For the TPAVI with ResNet baseline method, our method has significant performance gains across all metrics on both subsets.
On S4 subset, our model achieves 5.36 $ \mathcal{M}_{\mathcal{J}}$ (mIoU) gains and reaches 0.887 $ \mathcal{M}_{\mathcal{F}} $, which advances the baseline model with 0.848 $ \mathcal{M}_{\mathcal{F}} $ by a large margin.
On the more challenging MS3 subset, our model achieves higher gains of 6.20 on $ \mathcal{M}_{\mathcal{J}} $.
As for the TPAVI with PVT, although the performance of the baseline is very strong, our method can also bring performance gains.
For instance, our method achieves 3.75 $\mathcal{M}_{\mathcal{J}}$ gains on S4 subset and 5.49 $\mathcal{M}_{\mathcal{J}}$ gains on MS3 subset.
In terms of the AVSegFormer baseline method, even it is already a very powerful method, our method can also improve the performance upon it on both backbones.
For the ResNet backbone, our method achieves 2.51 $\mathcal{M}_{\mathcal{J}}$ gains on S4 subset and 6.35 $\mathcal{M}_{\mathcal{J}}$ gains on MS3 subset.
With PVT backbone, our method achieves new state-of-the-art performance: 83.15 $\mathcal{M}_{\mathcal{J}}$ on S4 subset and 61.95 $\mathcal{M}_{\mathcal{J}}$ on MS3 subset.

\begin{table}
\setlength{\tabcolsep}{3.4pt}
\small
 \centering
\begin{tabular}{ccccccccc}
\toprule
\multirow{2}{*}{Method} & \multirow{2}{*}{I.B.} & \multicolumn{2}{c}{S4} & \multicolumn{2}{c}{MS3}  \\
\cmidrule(lr){3-4} \cmidrule(lr){5-6} 
& & $ \mathcal{M}_{\mathcal{J}} $ & $ \mathcal{M}_{\mathcal{F}} $ & $\mathcal{M}_{\mathcal{J}}$ & $\mathcal{M}_{\mathcal{F}}$ \\
\midrule
$\text{TPAVI~\cite{zhou2022avs}}$  & ResNet & 72.80 & .848 & 47.90 & .578  \\
\text{{\scriptsize (ECCV'22)}} & PVT & 78.70 & .879 & 54.00 & .645 \\
\midrule 
$\text{ECMVAE~\cite{Mao_2023_ICCV}}$ & ResNet & 76.33 & .865 & 48.69 & .607 \\
\text{{\scriptsize (ICCV'23)}}& PVT& 81.74 & .901 & 57.84 & \underline{.708}\\
\midrule 
$\text{CATR~\cite{li2023catr}}$ & ResNet & 74.80 & .866 & 52.80 & .653\\
\text{{\scriptsize (ACMMM'23)}}   & PVT & 81.40 & .896 & 59.00 & .700 \\
\midrule 
{$\text{AVSC~\cite{liu2023audio1}}$} &ResNet  & 77.02 & .852 & 49.58 & .615  \\
\text{{\scriptsize (ACMMM'23)}}  & PVT& 80.57 & .882 & 58.22 & .651 \\
\midrule
{$\text{AuTR~\cite{liu2023audio}}$}   & ResNet & 75.00 & .852 & 49.40 & .612\\
\text{{\scriptsize (arXiv'23)}}  & PVT& 80.40 & .891 & 56.20 & .672 \\
\midrule 
SAMA-AVS~\cite{liu2024annotation}~\text{{\scriptsize (WACV'24)}}~  & ViT-H& 81.53 & .886 & \textbf{63.14} & .691 \\
\midrule 
{$\text{AVSegFormer~\cite{gao2023avsegformer}}$} & ResNet & 76.45 & .859 & 49.53 & .628\\
\text{{\scriptsize (AAAI'24)}}  & PVT& 82.06 & .899 & 58.36 & .693 \\
\midrule 
\midrule 
{\textbf{Ours} }   &ResNet & 78.15 & .887 & 54.08 & .616\\
{\scriptsize (w/ TPAVI)}& PVT & \underline{82.49} & \textbf{.912} & {59.49} & .676 \\
\midrule 
{\textbf{Ours} }   &ResNet & 78.96 & .875 & 55.88 & .645 \\
{\scriptsize (w/ AVSegFormer)}& PVT & \textbf{83.15} &  \underline{.904} & \underline{61.95} & \textbf{.709}   \\
\bottomrule
\end{tabular}
\caption{Comparison with up-to-date state-of-the-arts on both subsets. Our proposed methods significantly improve
the competitiveness of the baseline models. (The best performance in \textbf{bold} and the second best is \underline{underlined}; ``I.B.'' denotes image backbone.)
}
\label{tab:otherarts}
\end{table}

\noindent \textbf{Comparison with Other Arts.}
We also collect up-to-date AVS methods  AVSC~\cite{liu2023audio1}, CATR~\cite{li2023catr}, AuTR~\cite{liu2023audio}, CMVAE~\cite{Mao_2023_ICCV}, and SAMA-AVS~\cite{liu2024annotation}; and we compare the performance of these methods with our proposed method.
The results are shown in~\cref{tab:otherarts}.
The comparison shows the strong competitiveness of our proposed framework when compared with so various latest methods.
Our method based on AVSegFormer~\cite{gao2023avsegformer} is still the state-of-the-art method among all the methods.
Moreover, on S4 subset, the original TPAVI method only has 72.8 $\mathcal{M}_{\mathcal{J}}$ with ResNet, 78.7 $\mathcal{M}_{\mathcal{J}}$ with PVT, which falls behind the other methods including AVSC~\cite{liu2023audio1}, CATR~\cite{li2023catr}, AuTR~\cite{liu2023audio}, ECMVAE~\cite{Mao_2023_ICCV}.
However, by combining our method with the TPAVI, the model (Ours w/TPAVI) has outperformed the other methods including AVSegFormer, AVSC, CATR, AuTR and ECMVAE in almost all metrics; and it becomes the second best model except our AVSegFormer-based model ``ours w/ AVSegFormer'' under the same backbones.
The results clearly reveal the effectiveness of our versatile proposed framework.
Notably, our framework can also be applied on these methods~\cite{liu2023audio1,li2023catr,liu2023audio,Mao_2023_ICCV,liu2024annotation}  to further improve their performance.

\begin{table}[]
\small
\begin{center}
\begin{tabular}{cccccc}
\toprule
\multirow{2}{*}{Method}  & \multirow{2}{*}{I.B.}  & \multicolumn{2}{c} {5\%} & \multicolumn{2}{c}{10\%}    \\
\cmidrule(lr){3-4}\cmidrule(lr){5-6} 
 &  & $\mathcal{M}_{\mathcal{J}}$ & $\mathcal{M}_{\mathcal{F}}$  & $\mathcal{M}_{\mathcal{J}}$ & $\mathcal{M}_{\mathcal{F}}$  \\
 \midrule
\multirow{2}{*}{TPAVI} & ResNet & 57.34  & .733 & 57.91 & .746 \\
& PVT  & 67.06 & .794  &  71.72 & .830  \\
\midrule
{\textbf{Ours}} & ResNet  & 61.74 & .770 & 66.67 & .820 \\
 {\scriptsize (w/ TPAVI)} & PVT &  72.96 & .848 & 76.23 & .871  \\
\midrule
\midrule
\multirow{2}{*}{AVSegFormer} & ResNet  & 56.13 & .703  & 62.47 & .754 \\
& PVT & 68.35 & .797 & 73.92 & .840  \\
\midrule
{\textbf{Ours} } & ResNet & 64.96  & .766 & 69.26 & .796  \\
{\scriptsize (w/ AVSegFormer)} & PVT & 75.38 & .846 & 77.40 & .864 \\
\bottomrule
\end{tabular}
\end{center}
\vspace{-0.55cm}
\caption{Resutls of the models with different percentages of labeled training data from AVSBench S4 dataset.}
\label{tab:s4partial}
\end{table}

\noindent \textbf{Results using less labeled training data.}
We also investigate the performance when utilizing training data with varying proportions (5\% and 10\%) of labeled data on the S4 subset.
As shown in~\cref{tab:s4partial}, our approach consistently demonstrates impressive improvements across different data proportions. 
For instance, using only 10\% of labeled data with ResNet backbone, the performance of our model increases by nearly 10 points on $\mathcal{M}_{\mathcal{J}}$.
This indicates that our method is also highly effective when the labeled data is limited.

\noindent \textbf{Pre-training on the Single-source subset.} Following the TPAVI~\cite{zhou2022avs}, we conduct an investigation into the impact on the MS3 when using pretrained weights of S4. 
As shown in Table~\ref{tab:pretrain}, it is evident that pretraining on the S4 dataset will indeed
improve the performance on the MS3 subset for all methods including ours; and our method achieves the best performance on the MS3 subset among all methods using S4-pretrained weights.

\begin{table}
\small
\centering
\begin{tabular}{ccccc}
\toprule
\multirow{2}{*}{Method} & \multicolumn{2}{c}{From Scratch} & \multicolumn{2}{c}{P.T. on S4} \\
\cmidrule(lr){2-3} \cmidrule(lr){4-5}
& ResNet & PVT & ResNet & PVT \\
\midrule
TPAVI & 47.90 & 54.00 & 54.30 & 57.30 \\
\textbf{Ours} {\scriptsize (w/ TPAVI)}  & \textbf{54.08} & \textbf{59.49} & \textbf{54.73} & \textbf{60.78} \\

\midrule
AVSegFormer  & 49.53 & 58.36 & 55.78 & 61.91 \\

\textbf{Ours} {\scriptsize (w/ AVSegFormer)} & \textbf{55.88} & \textbf{61.95} & \textbf{59.32} & \textbf{64.47} \\ 
\bottomrule
\end{tabular}
\caption{Performance with different initialization strategies under the MS3 setting on $ \mathcal{M}_{\mathcal{J}} $.}
\label{tab:pretrain}
\end{table}

\subsection{Ablation Study}
In this section, we conduct ablation studies to evaluate the components in the framework with TPAVI baseline model on the S4 subset.

\begin{table}
    \begin{subtable}{0.5\linewidth}
    \scriptsize

        \centering
        
       \begin{tabular}{c c|c c}
\hline
 \textbf{NF} & \textbf{DF} & $\mathcal{M}_{\mathcal{J}}$ & $\mathcal{M}_{\mathcal{F}}$ \\
 \hline
 \ding{56} & \ding{56} & 72.80 & .848 \\
\ding{52} & \ding{56} & 76.17 & .869  \\
\ding{56} & \ding{52} & 77.43 & .881  \\
\ding{52} &\ding{52} & 78.15 & .887 \\
\hline
\end{tabular}
        \caption{Results with ResNet backbone.}
        
    \end{subtable}%
    \begin{subtable}{0.5\linewidth}
    \scriptsize

        \centering
        
        \begin{tabular}{c c|c c}
\hline
 \textbf{NF} & \textbf{DF} & $\mathcal{M}_{\mathcal{J}}$ & $\mathcal{M}_{\mathcal{F}}$ \\
 \hline
 \ding{56} & \ding{56} & 77.89 & .880 \\
\ding{52} & \ding{56} & 81.01 & .903 \\
\ding{56} & \ding{52} & 81.41 &   .905 \\
\ding{52} &\ding{52} & 82.49 & .912 \\
\hline
\end{tabular}
        
        \caption{Results with PVT backbone.}
    \end{subtable}
    \caption{Ablation study of our framework based on TPAVI baseline model on S4 subset.}
    \label{ab:nfdf}
\end{table}

\begin{table}
\setlength{\tabcolsep}{2.5pt}

    \begin{subtable}{0.5\linewidth}
    \scriptsize

        \centering
                \begin{tabular}{ccccc}
\toprule
\textit{Burn-in} & 5 & 10 & 20 &   30  \\
\midrule
$\mathcal{M}_{\mathcal{J}}$ & 77.72   &  \textbf{78.15}    & 77.50  &  77.48\\
$\mathcal{M}_{\mathcal{F}}$ &  0.883   & \textbf{0.887}    & 0.881 &   0.882\\
\bottomrule
\end{tabular}

        \caption{Burn-in epochs for training.}
        \label{tab:ablation-burnin}
        
    \end{subtable}%
    \begin{subtable}{0.5\linewidth}
    \scriptsize

        \centering
        
       \begin{tabular}{ccccc}
\toprule
$\lambda$  & 0.1 & 0.2 & 0.5 &   1.0  \\
\midrule
$\mathcal{M}_{\mathcal{J}}$ & 77.00   &   77.62  &  \textbf{78.15}  & 77.19 \\
$\mathcal{M}_{\mathcal{F}}$ &  0.878   & 0.883    & \textbf{0.887} & 0.881  \\
\bottomrule
\end{tabular}
        
        \caption{Unsupervised loss weight $\lambda$.}
    \end{subtable}
    \caption{Effects of burn-in epochs and unsupervised loss weight.}
    \label{tab:hyper}
\end{table}

\begin{figure*}[!htb]
\centering
\vspace{0.1cm}
\includegraphics[width=.98\linewidth]{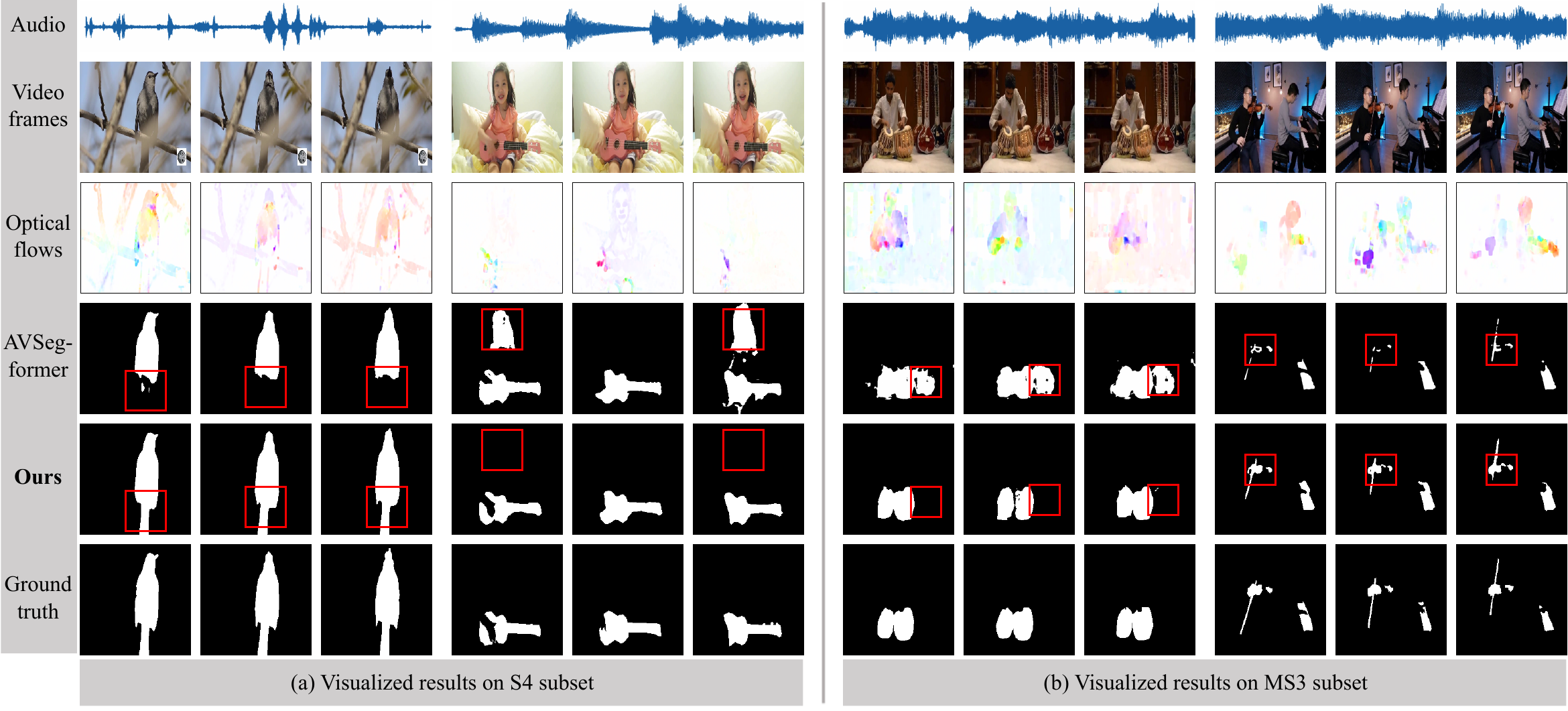}
\caption{Qualitative comparison between our method and AVSegFormer~\cite{gao2023avsegformer} on both subsets of AVSBench.
Our method shows better segmentation performance by localising the exact sounding object, attending to the fine-grained details and being closer to groundtruths.  
}
\label{fig:qual}
\end{figure*}

\noindent \textbf{Effectiveness of NF and DF.} To study the effects of neighboring frames (NFs) and distant frames (DFs) in our proposed framework, we conduct ablation studies, adjusting one component at a time based on the TPAVI baseline methods with two visual backbones. 
As illustrated in~\cref{ab:nfdf}, both NF and DF demonstrate significant enhancements when applied independently, indicating that their respective contributions to the performance are both fairly considerable. 
And when combining the NF and DF, the model obtains the best performance across all metrics.
The results demonstrate the effectiveness and complementarity of neighboring frames (NFs) and distant frames (DFs) in boosting the performance on the AVS task, implying the great value of the abundant unlabeled frames with proper exploitation.

\noindent \textbf{Epochs for burn-in stage.}
We ablate the burn-in epochs in the teacher-student training. 
As shown in~\cref{tab:hyper} (a), 10-epoch burn-in works best.
Either using the unsupervised loss too early or too late will result in a suboptimal performance.
If using the unsupervised loss too early, the pseudo labels from the teacher are not reliable thus it will cause negative effects on the final model performance.
If using the unsupervised loss is too late, the model will be biased toward the labeled data without utilizing the unlabeled data.

\noindent \textbf{Weight for the unsupervised loss.}
We ablate the weight $\lambda$ for the unsupervised loss $\mathcal{L}_\text{unsup}$.
The results in~\cref{tab:hyper} (b) show a moderate value of 0.5 achieving the best performance.
When $\lambda$ is set as low as 0.1, the improvement is less significant than the cases where $\lambda$ is 0.2 or 0.5.
However, if $\lambda$ is too high such as 1.0, the model performance degrades. 
This is due to the model overemphasizing the unlabeled data when using large unsupervised loss weight.

\subsection{Qualitative Examples}
In~\cref{fig:qual}, we qualitatively show some segmentation results of our method and the baseline AVSegFormer~\cite{gao2023avsegformer}.
The results clearly demonstrate the advantages of our method by producing the segmentations for the sounding objects which are closer to groundtruths.
As shown in the first video of~\cref{fig:qual} (a), with the assistance of flow, our method can segment the tail of the bird which is hard to find with only visual RGB frames.
In the second video of~\cref{fig:qual} (a), the AVSegFormer baseline falsely segments both the \textit{Ukulele} and the salient yet silent \textit{person}. 
On the contrary, our method accurately localizes and segments only \textit{Ukulele} according to the audio cues, without being distracted by the silent \textit{person}.
In the multi-sound scenario in~\cref{fig:qual} (b), our method also shows improved performance.
In the first video, the AVSegFormer~\cite{gao2023avsegformer} baseline mistakenly segments another silent instrument as marked with red boxes; while our method leverages both the sound and the hand motion of the player to produce the segmentations for the instrument being played.
In the second video, the segmentations produced by our method for both instruments have fine-grained details and are closer to groundtruth annotations.

\section{Conclusion}\label{sec:conclu}
In this paper, we have pointed out the major limitation of previous AVS methods: the \textit{underutilization} of the abundant unlabeled frames. 
To mitigate this, we analyzed that the unlabeled frames can be divided into two categories: \textit{neighboring frame} (NF) and \textit{distant frame} (DF), according to the temporal characteristics.
And we proposed a unified \textit{unlabeled frame exploitation} (UFE) framework to harness the two kinds of unlabeled frames based on their unique traits.
For NFs, we extracted the motion cues as dynamic guidance to assist in the precise localization of sounding objects; while for DFs, since they are natural semantic augmentations to the labeled frames, we utilized them to enrich the data diversity with the teacher-student training.
Extensive experiments have demonstrated the {significant} improvement brought by the exploitation of unlabeled frames.
We believe that our proposed framework serves as a strong baseline and hopefully inspires more research to value both labeled and unlabeled data to successfully tackle AVS task.


{
\small
\bibliographystyle{ieeenat_fullname}
\bibliography{refs}
}


\end{document}